\documentclass[letterpaper]{article} 
\usepackage{aaai25}  
\usepackage{times}  
\usepackage{helvet}  
\usepackage{courier}  
\usepackage[hyphens]{url}  
\usepackage{graphicx} 
\usepackage{natbib}  
\usepackage{caption} 
\usepackage{algorithm}
\usepackage{algorithmic}

\usepackage{amsmath}
\usepackage{amssymb}
\usepackage{multirow}
\usepackage{booktabs}
\usepackage{colortbl}
\usepackage{color,xcolor}
\definecolor{light-gray}{gray}{0.9}
\definecolor{light-gray0}{gray}{0.7}
\usepackage{array}
\usepackage{bm}
\usepackage{cases}
\usepackage{caption}

\usepackage{newfloat}
\usepackage{listings}
\DeclareCaptionStyle{ruled}{labelfont=normalfont,labelsep=colon,strut=off} 
\floatstyle{ruled}
\newfloat{listing}{tb}{lst}{}
\floatname{listing}{Listing}
\title{Promptable Anomaly Segmentation with SAM Through Self-Perception Tuning}
\author {
    Hui-Yue Yang\textsuperscript{\rm 1}, Hui Chen\textsuperscript{\rm 2}\thanks{Corresponding author.}, Ao Wang\textsuperscript{\rm 1}, Kai Chen\textsuperscript{\rm 1}, Zijia Lin\textsuperscript{\rm 1}, \\ Yongliang Tang\textsuperscript{\rm 3}, Pengcheng Gao\textsuperscript{\rm 3}, Yuming Quan\textsuperscript{\rm 3}, Jungong Han\textsuperscript{\rm 4}, Guiguang Ding\textsuperscript{\rm 1}
}
\affiliations {
    \textsuperscript{\rm 1}School of Software, Tsinghua University \\
    \textsuperscript{\rm 2}BNRist, Tsinghua University \\
    \textsuperscript{\rm 3}LUSTER LightTech Co., Ltd.\\
    \textsuperscript{\rm 4}Department of Automation, Tsinghua University \\

    yanghuiyue18@outlook.com, 
    \{jichenhui2012,chenkai2010.9,quanym,jungonghan77\}@gmail.com, 
    wa22@mails.tsinghua.edu.cn, 
    linzijia07@tsinghua.org.cn, 
    yongliangtang@lusterinc.com, 
    gaopengcheng15@mails.ucas.ac.cn, 
    dinggg@tsinghua.edu.cn 
}

\begin{document}

\maketitle

\begin{abstract}
Segment Anything Model (SAM) has made great progress in anomaly segmentation tasks due to its impressive generalization ability. However, existing methods that directly apply SAM through prompting often overlook the domain shift issue, where SAM performs well on natural images but struggles in industrial scenarios. Parameter-Efficient Fine-Tuning (PEFT) offers a promising solution, but it may yield suboptimal performance by not adequately addressing the perception challenges during adaptation to anomaly images. In this paper, we propose a novel \textbf{S}elf-\textbf{P}erceptinon \textbf{T}uning (\textbf{SPT}) method, aiming to enhance SAM's perception capability for anomaly segmentation. The SPT method incorporates a self-drafting tuning strategy, which generates an initial coarse draft of the anomaly mask, followed by a refinement process. Additionally, a visual-relation-aware adapter is introduced to improve the perception of discriminative relational information for mask generation. Extensive experimental results on several benchmark datasets demonstrate that our SPT method can significantly outperform baseline methods, validating its effectiveness.
\end{abstract}

\begin{links}
    \link{Code}{https://github.com/THU-MIG/SAM-SPT}
    \link{Extended version}{https://arxiv.org/pdf/2411.17217}
\end{links}

\section{Introduction}

Anomaly segmentation aims to automatically locate and segment abnormal regions within images of industrial products, which is crucial for improving production efficiency and product quality \cite{destseg,clipad,ras}. In real-world scenarios, the sheer volume of industrial products and the wide spectrum of anomaly types present significant challenges for anomaly segmentation tasks. Consequently, constructing a highly generalized anomaly segmentation model has become a focal point of interest in the field \cite{saa,winclip}.

\begin{figure}[!t]
\centering
\includegraphics[width=\linewidth]{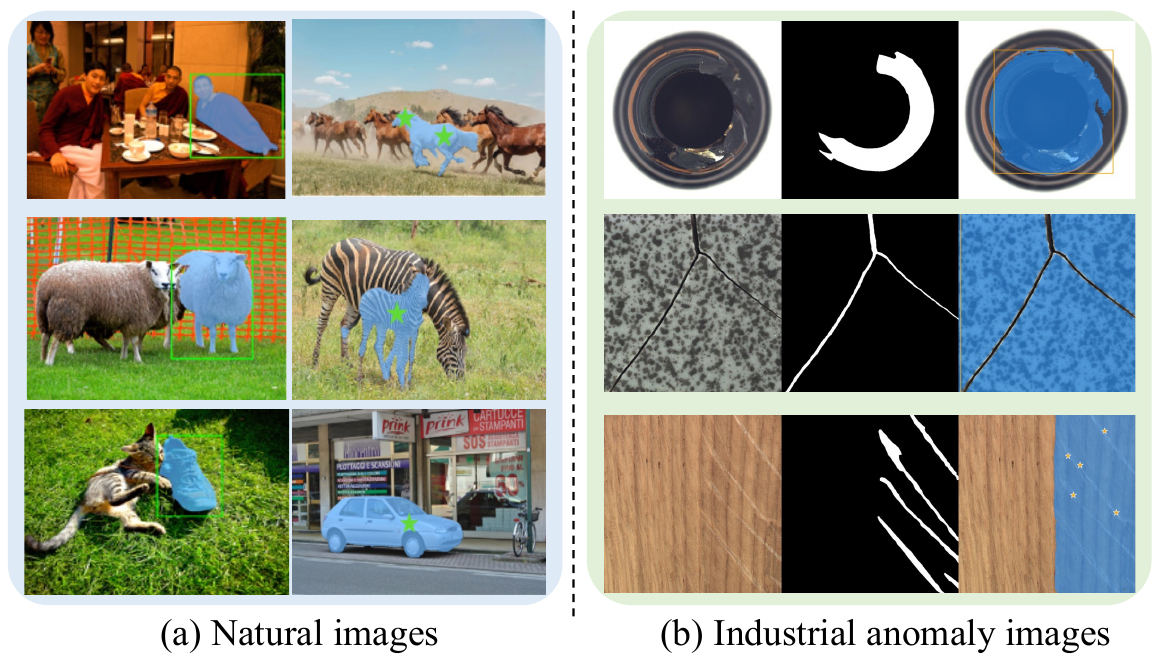}
\caption{Illustration of the domain shift issue. SAM performs well on natural images but poorly on out-of-domain industrial anomaly images.} 
\label{fig:domainshift}
\end{figure}

Recently, thanks to the emergence of pivotal vision foundation models like the Segment Anything Model (SAM) \cite{sam}, significant advancements have been made in various computer vision fields. SAM, pre-trained on an extensive dataset SA-1B \cite{sam}, has showcased remarkable zero-shot visual perception capabilities through prompting \cite{seem,vrp-sam,hq-sam, repvit, repvitsam}. Consequently, extensive research is underway to leverage SAM's robust generalization abilities for advancing anomaly segmentation tasks.

For example, SAA \cite{saa} represents a pioneering effort that applies SAM to anomaly segmentation tasks. Initially, it utilizes prompt-guided object detection methods like GroundingDINO \cite{dino} to generate prompt-conditioned box-level regions that highlight desired anomaly areas. These boxes are then fed into SAM as prompts to generate final predictions for anomaly segmentation. SAA+ \cite{saa} further introduces hybrid prompts that incorporate domain-specific expertise with contextual image information to mitigate ambiguities inherent in language prompts. In contrast, CLIPSAM \cite{clipsam} replaces GroundingDINO with CLIP \cite{clip}, leveraging its enhanced capability to provide precise localization information as prompts for SAM.

\begin{figure}[ht]
\centering
\includegraphics[width=\linewidth]{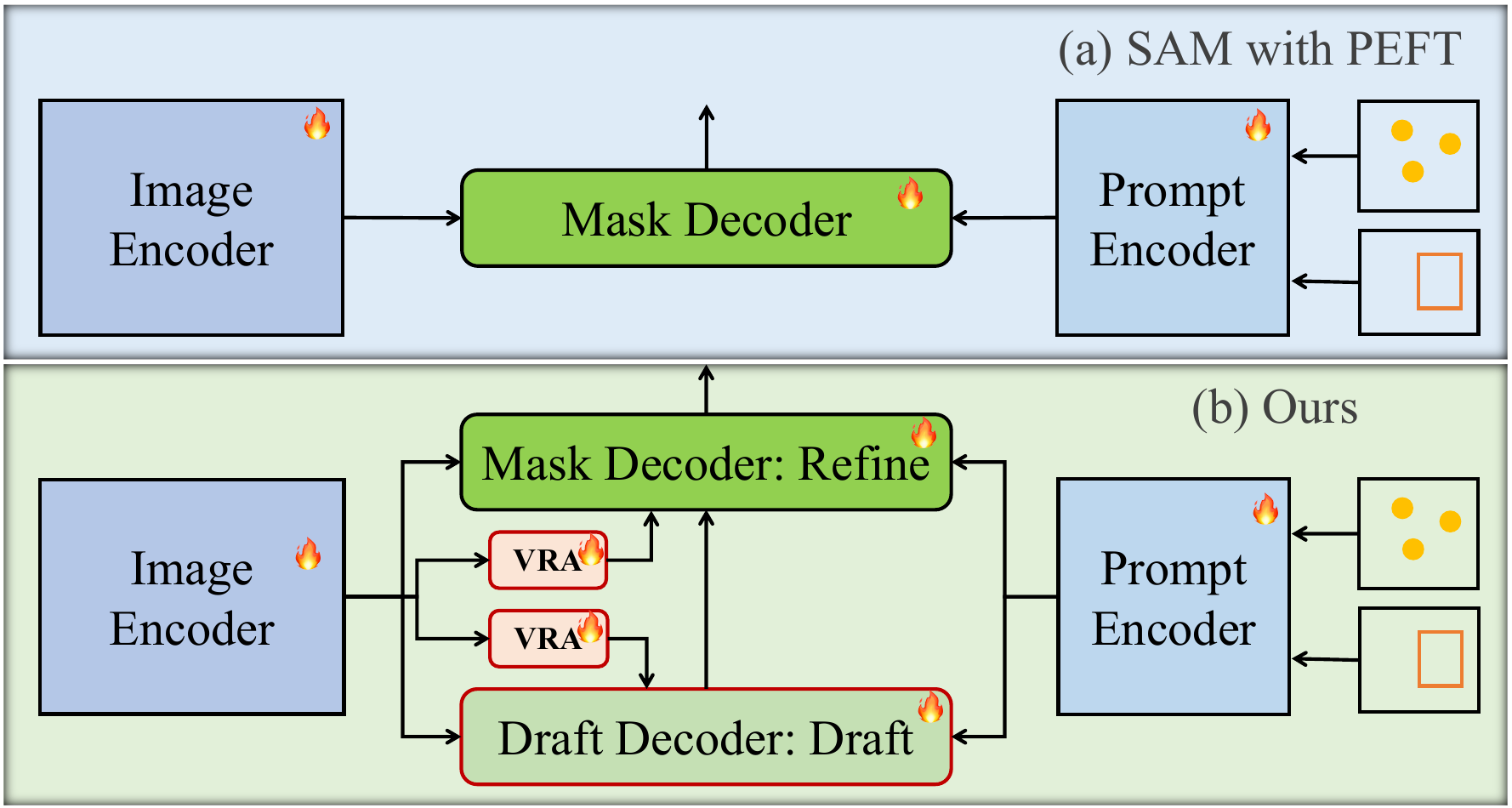}
\caption{A comparison between SAM with PEFT methods and our promptable anomaly segmentation model with self-perception tuning. VRA denotes the VRA-Adapter.} 
\label{fig:modelcomparison}
\end{figure}

Despite the remarkable progress achieved, existing methods often overlook a critical issue of domain shift \cite{dasurvey,c-visdit} when adapting SAM to downstream tasks \cite{zhang2024improving,chen2023sam,huang2023robustness}. Specifically, SAM is primarily trained on natural image datasets, which enables excellent performance in general visual understanding tasks. However, it typically suffers from significant degradation when applied to out-of-domain industrial scenarios (see Fig. \ref{fig:domainshift}). Therefore, existing methods compensates for SAM's limited perception of domain-specific distributions by offering more contextual prompts. However, the fundamental issue of domain shift is still insufficiently considered in the context of anomaly segmentation.

Fine-tuning is a straightforward approach to enhance SAM's adaptation in anomaly segmentation tasks. Current research often employs parameter-efficient fine-tuning (PEFT) methods like LoRA \cite{lora} and Adapter \cite{adapter} to tune foundation models. These methods introduce a minimal number of trainable parameters while keeping the bulk of SAM's parameters frozen. Despite notable improvements observed in our experiments, simply applying traditional PEFT methods to anomaly segmentation may obtain suboptimal results without taking the task challenge into consideration. Particularly, diverse imaging conditions, varied anomaly types across different industrial products, and the complex appearances of anomalies can greatly hinder the perception ability during adaptation. These challenges hinder the robustness of segmentation across different prompts as observed in our experiments (Fig. \ref{fig7:vis_box_point}). Therefore, a question is raised: \textit{How to enhance the perception ability of SAM to broaden its generalization capability for anomaly segmentation?}

In this paper, we propose a promptable anomaly segmentation model with SAM through a novel \textbf{S}elf-\textbf{P}erception \textbf{T}uning (\textbf{SPT}) method. Compared with conventional PEFT methods, the proposed SPT enhances perception capabilities by leveraging both \textbf{external} knowledge and \textbf{internal} priors, both of which are inherited from the model itself. Specifically, as shown in Fig. \ref{fig:modelcomparison}, we firstly design a Self-Draft Tuning (SDT) strategy, in which SAM initially generate a coarse draft of the anomaly mask, followed by a mask refinement process. This draft can serve as an external yet coarse perception knowledge of the anomaly mask, boosting the mask decoder with additional priors beyond the provided prompts and learned visual patterns. Furthermore, we introduce a Visual-Relation-Aware Adapter (VRA-Adapter) to enhance the internal perception knowledge of discriminative relation information during decoding. By capturing the relationship among different regions, our adapter can enable decoders to achieve a more nuanced and detailed understanding of anomaly patterns, resulting in more consistent and accurate anomaly segmentation. We show that our method can accommodate different PEFT methods and offer a flexible tuning framework. Experimental results on several industrial datasets show that compared to baseline methods, our SPT can enhance SAM's robust generalization for anomaly segmentation under different prompts, well demonstrating its effectiveness and superiority.

To summarize, our contributions are as follows:
\begin{itemize}
\item We introduce a promptable anomaly segmentation model with SAM through a novel Self-Perception Tuning (SPT) method. SPT can improve the perception capability of anomalies, enhancing the robust generalization of SAM for anomaly segmentation under various prompts.

\item We design a self-draft tuning strategy to boost the mask decoder with external yet coarse perception of anomaly mask. A visual-relation-aware adapter is further introduced to enhance the internal perception of discriminative relation information for anomaly mask decoding. 

\item Extensive experiments across various industrial datasets for anomaly segmentation tasks show that our method can achieve state-of-the-art performance, well demonstrating the effectiveness of our method.
\end{itemize}

\section{Related Work}
\textbf{Foundation models for anomaly segmentation. }
Large pre-trained models like CLIP and SAM have significantly advanced anomaly segmentation. WinCLIP~\cite{winclip} first leverages CLIP's vision-language representation for zero-shot and few-shot anomaly detection. Clip-AD~\cite{clipad} refines CLIP for better zero-shot detection of unseen anomalies, while AnomalyCLIP~\cite{anomalyclip} utilizes CLIP's pre-training for effective anomaly detection without extensive task-specific training. SAA~\cite{saa} uses GroundingDINO and SAM for segmentation and designs rules to filter out anomalies that meet specific criteria. CLIPSAM~\cite{clipsam} integrates CLIP's semantic understanding with SAM's segmentation capabilities, enhancing anomaly detection in complex scenes. These methods collectively demonstrate the effectiveness of integrating large pre-trained models into anomaly detection frameworks. However, they generally directly apply the foundation models into anomaly segmentation tasks, neglecting the critical issue of domain shift between pre-trained datasets and downstream tasks. In contrast, our work aims to construct a robust SAM with great generalization for anomaly segmentation tasks through fine-tuning strategy, which is complementary to the existing methods. 

\begin{figure*}[ht]
\centering
\includegraphics[width=\linewidth]{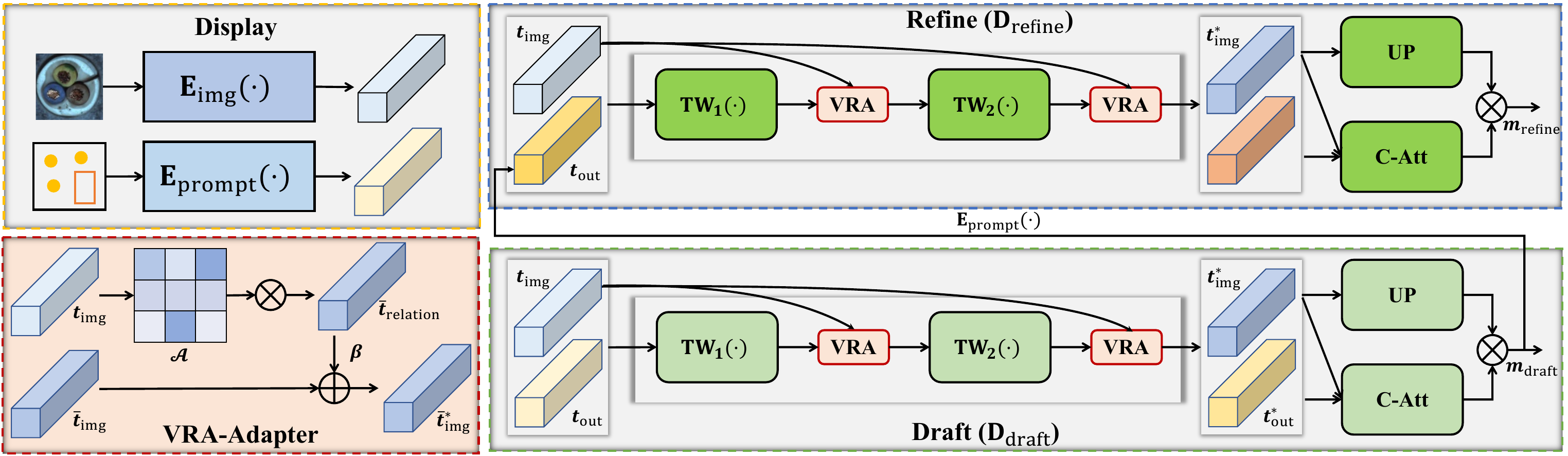}
\caption{Overview of the proposed Self-Perception Tuning (SPT) framework, which applies a self-draft tuning (SDT) strategy and visual-relation-aware adapters (VRA-Adapter) to enhance the perception ability of SAM. SDT consists of three phases, \textit{i.e.,} display, draft, and refine. VRA is an abbreviation for VRA-Adapter.} 
\label{fig:framework}
\end{figure*}

\textbf{Parameter-efficient fine-tuning (PEFT).} These methods address the high computational costs of adapting large pre-trained models into downstream tasks \cite{lora,pyra,adapter,consolidator}. Adapter \cite{adapter}  introduces a small amount of trainable layers to each pre-trained layer, enabling efficient task-specific adaptation. LoRA \cite{lora} further reduces trainable parameters by injecting low-rank matrices into the model's layers. Adaptformer\cite{adaptformer} introduces lightweight modules that add minimal parameters to the pre-trained ViTs. SamAdapter~\cite{samadapter} applies PEFT to SAM models, improving performance in underperforming scenes and challenging tasks. Recent advances in PEFT methods include DoRA~\cite{dora} which introduces a weight-decomposed low-rank adaptation approach for efficient fine-tuning. NOLA~\cite{nola} compresses LoRA using a linear combination of random basis for more efficient training. These methods demonstrate the ongoing evolution and versatility of PEFT techniques in the field of model adaptation. Despite notable improvement observed in our experiments, simply applying these PEFT methods to anomaly segmentation may take no consideration of the perception difficulties during adaptation, leading to suboptimal results. Therefore, we propose a self-perception tuning method to enhance the perception ability of SAM. As illustrated in our experiments, our SPT can accommodate different traditional PEFT methods, offering a flexible and effective fine-tuning framework.

\section{Methodology}
Here, we introduce a novel self-perception tuning (SPT) method to adapt SAM into the anomaly segmentation tasks with enhanced perception ability, as illustrated in Fig. \ref{fig:framework}.

\subsection{Preliminary}\label{sec3.1}

\paragraph{SAM}
The Segment Anything Model (SAM) is an interactive image segmentation model based on user-provided prompts such as points, boxes, or text. It consists of an image encoder, a prompt encoder, and a mask decoder. The image encoder first extracts the image feature representation by a Vision Transformer (ViT) \cite{dosovitskiy2020image}. Then, SAM uses the prompt encoder to represent the user-provided prompts with prompt embeddings. Finally, the mask decoder utilizes the image features and prompt embeddings to generate the segmentation mask for the target object.

SAM aims to achieve high flexibility and generality, making it adaptable to various segmentation tasks without the need for task-specific fine-tuning. However, due to the domain shift between pre-trained datasets and practical industrial scenarios, directly applying SAM for anomaly segmentation unavoidably encounters performance degradation. 

\subsection{Self-Draft Tuning}\label{sec3.2}
Directly applying conventional PEFT methods (\textit{e.g.,} LoRA) to SAM for anomaly segmentation tasks may lead to suboptimal performance, as these methods do not account for the task challenge to the perception ability of SAM. To mitigate this issue, the proposed SPT method firstly introduces a self-draft tuning (SDT) strategy to enhance the perception outcome of SAM by utilizing the external knowledge of the model itself. In SDT, we augment SAM with a draft decoder alongside its the original mask decoder. When given an image, SAM firstly generates a coarse draft of the anomaly mask using the draft decoder. Subsequently, the mask decoder refines this draft, resulting in more accurate segmentation. For clarity, we organize the SDT pipeline into three stages: display, draft, and refine.

\textbf{Display.} In the display phase, SAM uses the image encoder to extract the image features $\bm{e}_{\text{img}}$ for the input image $\bm{I}$ and uses the prompt encoder to generate various types of prompt embeddings for user-provided prompts $\bm{P}$, including sparse embedding $\bm{e}_{\text{sparse}}$, dense embedding $\bm{e}_{\text{dense}}$, and position embedding $\bm{e}_{\text{pos}}$:
\begin{equation}
\label{eq: display}
    \bm{e}_{\text{img}} = \mathbf{E}_{\text{img}}(\bm{I}), \{\bm{e}_{\text{sparse}}, \bm{e}_{\text{dense}}, \bm{e}_{\text{pos}}\} = \mathbf{E}_{\text{prompt}}(\bm{P})
\end{equation}
where $\mathbf{E}_{\text{img}}$ denotes the image encoder and $\mathbf{E}_{\text{prompt}}$ denotes the prompt encoder. We augment both encoders with learnable PEFT modules to capture discriminative feature representations for anomaly segmentation.

\textbf{Draft.} The draft decoder shares the same structure as the original mask decoder and is initialized with pre-trained weights of the latter, maintaining considerable segmentation ability for draft. It utilizes separated Two-Way Transformer blocks to manage the input information including the image feature $\bm{e}_{\text{img}}$, the prompt embeddings $\{\bm{e}_{\text{sparse}}, \bm{e}_{\text{dense}}, \bm{e}_{\text{pos}}\}$, the IoU token $\bm{e}_{\text{iou}}$ and mask tokens $\bm{e}_{\text{mask}}$:
\begin{equation}
\begin{aligned}
\label{equ:draft_tw}
\bm{t}_{\text{img}}&=\bm{e}_{\text{img}} + \bm{e}_{\text{dense}} \\
\bm{t}_{\text{out}}&=[\bm{e}_{\text{sparse}}, \bm{e}_{\text{iou}}, \bm{e}_{\text{mask}}] \\
\bm{t}_{\text{img}}^*, \bm{t}_{\text{out}}^*&=\mathbf{TW}_2(\mathbf{TW}_1(\bm{t}_{\text{img}}, \bm{e}_{\text{pos}}, \bm{t}_{\text{out}}))
\end{aligned}
\end{equation}
where $\mathbf{TW}(\cdot)$ with subscripts is the process of the Two-Way Transformer Block. $[\cdot]$ means token concatenation.
The final mask can be generated by a dot product between the updated mask token and the upsampled image features:
\begin{equation}
\label{equ:draft_gen}
\bm{e}_{\text{mask}}^* = \mathbf{\text{C-Att}}(\bm{t}_{\text{out}}^*, \bm{t}_{\text{img}}^*)[-1], \bm{m}_{\text{draft}} = \bm{e}_{\text{mask}}^* \times \text{UP}(\bm{t}_{\text{img}}^*)
\end{equation}
where $\mathbf{\text{C-Att}}(\cdot, \cdot)$ is a token-to-image cross attention. The original SAM has 3 mask tokens. Here, we only use the first token for mask generation by default following previous works \cite{hq-sam}.
Combining Eq. \ref{equ:draft_tw} and Eq. \ref{equ:draft_gen}, the coarse draft generation can be formalized as follows:
\begin{equation}
\label{eq: d_draft}
    \bm{m}_{\text{draft}} = \mathbf{D}_{\text{draft}}(\bm{e}_{\text{img}}, \bm{e}_{\text{sparse}}, \bm{e}_{\text{dense}}, \bm{e}_{\text{pos}})
\end{equation}
where $\mathbf{D}_{\text{draft}}$ denotes the draft decoder. We use PEFT methods to adapt the draft decoder to the anomaly segmentation.

\textbf{Refine.} In the refine phase, the original mask decoder of SAM aims to perceive all available information to generate a refined mask for prediction. Therefore, we derive the representation for the coarse draft of anomaly mask, \textit{i.e.,} $\bm{m}_{\text{draft}}$, through the prompt encoder $\mathbf{E}_\text{prompt}$:
\begin{equation}
\bm{e}_{\text{draft}} = \mathbf{E}_\text{prompt}(\bm{m}_{\text{draft}})
\end{equation}
Noting that $\bm{e}_{\text{draft}}$ captures the coarse perception of the anomaly mask for SAM. Consequently, we can use it as a external knowledge to refine the mask generation for the anomaly segmentation. The final mask for anomaly segmentation is derived by:
\begin{equation}
    \bm{m}_{\text{refine}} = \mathbf{D}_{\text{refine}}(\bm{e}_{\text{img}}, \bm{e}_{\text{sparse}}, \bm{e}_{\text{draft}}, \bm{e}_{\text{pos}})
\end{equation}
where $\mathbf{D}_{\text{refine}}$ follows the same procedure as $\mathbf{D}_{\text{draft}}$ in Eq. \ref{eq: d_draft} but using the original mask decoder of SAM. $\bm{e}_{\text{draft}}$ is injected as the dense embedding. The PEFT method is applied to adjust the mask decoder to refine the anomaly mask.

By reviewing the SDT framework, we observe that SAM’s feature representations for images and prompts are tailored to industrial scenarios during the display phase. Subsequently, the decoders are fine-tuned to align with the anomaly generation task through a draft-then-refine procedure, effectively enhancing SAM's perception capability for anomaly segmentation.

\subsection{Visual-Relation-Aware Adapter}\label{sec3.3}
The Visual-Relation-Aware Adapter (VRA-Adapter) aims to enhance the mask decoders by integrating the visual relationships among different regions into the decoding process. It starts by evaluating visual relationships within the image and then uses a relation-aware adapter to propagate this relation knowledge to the decoder.

\textbf{Visual relation evaluation} is based on the insight that inter-region relationships within an image provide valuable information for anomaly detection \cite{you2022unified,yao2023focus}. Therefore, we firstly measure such relationships using the cosine similarity metric:
\begin{equation}
\mathbf{S}=\text{cosine}(\bm{e}_{\text{img}}, \bm{e}_{\text{img}})
\end{equation}
where $\bm{e}_{\text{img}}$ is the feature embeddings corresponding to each image region derived in Eq. \ref{eq: display}. The diagonal elements of $\mathbf{S}$ is set to $-\infty$ to avoid trivial self-similarities. Subsequently, we apply a softmax function to obtain the relation matrix:
\begin{equation}
\mathbf{\mathcal{A}} = \text{softmax}(\mathbf{S})
\end{equation}

The relation matrix is further refined using a thresholding mechanism controlled by a parameter $\alpha$ divided by the feature dimensionality $d$, ensuring that only the most significant relations are retained:
\begin{equation}
\label{eq:relationmatrix}
\mathbf{\mathcal{A}}^* = \begin{cases} 
\mathbf{\mathcal{A}}_{ij} & \text{if } \mathbf{\mathcal{A}}_{ij} \geq \frac{\alpha}{d} \\
0 & \text{otherwise}
\end{cases}
\end{equation}

\textbf{Relation-aware adapter} aims to integrate this relation information to enhance the quality of anomaly segmentation. Considering that neither decoder, \textit{i.e.,} $\mathbf{D}_{\text{draft}}$ and $\mathbf{D}_{\text{refine}}$, applies self-attention to the image embeddings $\bm{e}_{\text{img}}$, attention to image regions may gradually shift or even diminish over the course of decoding, leading to inconsistent anomaly segmentation. Therefore, we use a relation-aware adapter to complement the perception of relationship among image regions for the decoding process.

We denote the updated image features output by the Two-Way Transformers ($\mathbf{TW}(\cdot)$ in Eq. \ref{equ:draft_tw}) as $\bar{\bm{t}}_{\text{img}}^{1}$ and $\bar{\bm{t}}_{\text{img}}^{2}$. For ease of explanation, we use $\bar{\bm{t}}_{\text{img}}$ to refer to these features. Then, leveraging the relation matrix $\mathbf{\mathcal{A}}$, we can aggregate features with high relationships for each image feature via the attention mechanism, resulting in the relation-aware image features for decoding:
\begin{equation}
\bar{\bm{t}}_{\text{relation}} = \mathbf{\mathcal{A}}\bar{\bm{t}}_{\text{img}}
\end{equation}
Subsequently, we combine the relation-aware image feature with the origin image feature $\bar{\bm{e}}_{\text{img}}$ using a scale vector $\bm{\beta}$:
\begin{equation}
\bar{\bm{t}}^*_{\text{img}}=\bar{\bm{t}}_{\text{img}} + \bm{\beta}\bar{\bm{t}}_{\text{relation}}
\end{equation}
The scale vector $\bm{\beta}$ is dynamically adjusted during training to regulate the integration of visual relation information. 

By leveraging the internal perception knowledge of visual relationships, \textit{i.e.,} $\mathbf{\mathcal{A}}$, the image feature $\bar{\bm{t}}^*_{\text{img}}$ becomes more consistent and discriminative for regions with similar visual patterns. This enhancement significantly improves the discrimination between anomalous and normal regions, enabling a more accurate and detailed understanding of anomalies, and thus leading to better performance.

\subsection{Promptable Anomaly Segmentation Model}\label{sec3.4}

We fine-tune SAM using the proposed self-perception tuning method, resulting in a promptable anomaly segmentation model. The optimization objectives combine cross-entropy loss and Dice loss over both the draft and final refined masks:
\begin{equation}
\begin{aligned}
\mathcal{L} & = \mathcal{L}_{\text{CE}}(\mathbf{m}_{\text{draft}}, \mathbf{Y}) + \mathcal{L}_{\text{dice}}(\mathbf{m}_{\text{draft}}, \mathbf{Y}) \\
& + \mathcal{L}_{\text{CE}}(\mathbf{m}_{\text{refine}}, \mathbf{Y}) + \mathcal{L}_{\text{dice}}(\mathbf{m}_{\text{refine}}, \mathbf{Y}) \\
\end{aligned}
\end{equation}
where $\mathbf{Y}$ denotes the ground truth for the input image given the prompt.


\begin{table*}[!ht]
\belowrulesep=0pt
\aboverulesep=0pt
\renewcommand{\arraystretch}{1.1}
\centering
\small
\setlength{\tabcolsep}{1.5mm}
\caption{Performance comparison under different evaluation modes (\%). ``Avg.'' is the average scores of four kinds of prompts.}
\label{tab:vit_b}
\begin{tabular}{c|c|c|c|c|c|c|c|c|c|c}
\toprule
 & \multicolumn{2}{c|}{One box} & \multicolumn{2}{c|}{Multiple boxes} & \multicolumn{2}{c|}{Point=5} & \multicolumn{2}{c|}{Point=10} & \multicolumn{2}{c}{Avg.}\\
\cmidrule{2-11}
\multirow{-2}*{Method} & mIoU & mBIoU & mIoU & mBIoU & mIoU & mBIoU & mIoU & mBIoU & mIoU & mBIoU \\
\midrule
zero-shot & 56.3 & 50.7 & 63.0 & 57.9 & 44.1 & 39.8 & 44.4 & 39.7 & 52.0 & 47.0\\
\midrule
LoRA & 65.1 & 58.8 & 69.9 & 64.8 & 62.4 & 57.5 & 68.7 & 62.9 & 66.5 & 61.0 \\
\rowcolor{light-gray} SPT$_{\text{LoRA}}$ & \textbf{67.1} & \textbf{60.7} & \textbf{71.7} & \textbf{66.3} & \textbf{64.6} & \textbf{59.8} & \textbf{70.1} & \textbf{64.5} & \textbf{68.4} & \textbf{62.8}\\
\midrule
DoRA & 65.3 & 58.8 & 70.1 & 64.7 & 62.8 & 57.9 & 67.7 & 62.0 & 66.5 & 60.9 \\
\rowcolor{light-gray}  SPT$_{\text{DoRA}}$ & \textbf{66.9} & \textbf{60.6} & \textbf{71.4} & \textbf{66.4} & \textbf{65.3} & \textbf{60.4} & \textbf{70.5} & \textbf{64.9} & \textbf{68.5} & \textbf{63.1} \\
\midrule
Adapter & 65.1 & 59.2 & 70.0 & 65.3 & 61.8 & 56.9 & 67.2 & 61.4 & 66.0 & 60.7\\
\rowcolor{light-gray} SPT$_{\text{Adapter}}$ & \textbf{66.0} & \textbf{59.5} & \textbf{71.2} & \textbf{65.9} & \textbf{63.4} & \textbf{58.7} & \textbf{69.3} & \textbf{63.7} & \textbf{67.5} & \textbf{62.0} \\
\bottomrule
\end{tabular}
\end{table*}

\section{Experiments}
\subsection{Experiment Setups}
\paragraph{Datasets.} 
To ensure generalization across various industrial products and anomaly types with different prompts, we collect approximately 15,000 industrial anomaly images from real-world factories as training dataset. For the evaluation, we use six standard benchmark datasets, including MVTec \cite{mvtec}, VisA \cite{visa}, MTD \cite{mtd}, KSDD2 \cite{ksdd2}, BTAD \cite{btad}, and MPDD \cite{mpdd}. The training dataset includes a variety of imaging conditions, product types, and anomaly classes that are distinct from those in the test datasets, ensuring a fair evaluation of generalization. More details are provided in the extended version.

\paragraph{Implementation details.}
We initialize the SAM model with the officially provided pre-trained weights, utilizing three different backbone sizes: ViT-H, ViT-L, and ViT-B.
The input image is resized to $1024\times1024$. During training, for all models, the learning rate is set to $1\times 10^{-3}$ and is reduced after 10 epochs. All models are trained for 16 epochs using 8 NVIDIA 3090 GPUs with a batch of 8 images. The $\alpha$ in VRA-Adapter remains robust within the range of 0 to 0.5, depending on the specific PEFT method and model size. The rank of adapter is set to 8 for all models by default. 

\paragraph{Evaluation metrics.} 
To ensure that the constructed anomaly segmentation models can maintain the promptable functionality and generalization as SAM, we design three types of evaluation mode with different prompts: (1) one box, which highlights one or multiple defects with a single box; (2) multiple boxes, where each defect is assigned its own box, and (3) points. We randomly sample either 5 or 10 points for the prompt of points. We believe that this comprehensive evaluation effectively simulates user behavior in real-world scenarios, making our method highly applicable in practice. Two widely used metrics for promptable segmentation tasks, \textit{i.e.,} mean Intersection over Union (mIoU) and mean Boundary Intersection over Union (mBIoU), are leveraged to evaluate the segmentation result, following previous works \cite{sam,hq-sam}. 

\subsection{Comparison with State-of-the-art Methods}
We choose zero-shot SAM and representative PEFT methods as baseline methods, including LoRA~\cite{lora},  DoRA~\cite{dora}, and Adapter~\cite{adapter}. Since our method can accommodate the PEFT methods, we introduce three variants of our method according to its adopted PEFT method, having SPT$_{\text{LoRA}}$, SPT$_{\text{DoRA}}$, and SPT$_{\text{Adapter}}$. We provide comparison results based on ViT-B as backbone. More baseline methods and results of ViT-L/H are left in the extended version.

As illustrated in Table \ref{tab:vit_b}, our method significantly outperform various baseline approaches. Notably, compared to the zero-shot performance of the vanilla SAM model, our three SPT variants significantly enhance anomaly segmentation across six benchmark datasets, achieving over 15\% improvement in mIoU and mBIoU on average. This substantial improvement is due to our SPT method's effectiveness in addressing the domain shift issue for SAM. Additionally, compared to conventional PEFT baselines, our SPT method consistently delivers large performance gains. Specifically, SPT$_{\text{DoRA}}$ surpasses DoRA by an average of 2.0\% in mIoU and 2.2\% in mBIoU. In the Point=10 setting, these improvements increase to 2.8\% in mIoU and 2.9\% in mBIoU when comparing SPT$_{\text{DoRA}}$ to DoRA. These gains can be attributed to the proposed self-draft tuning framework and the visual-relation-aware adapter.

\subsection{Model Analysis}
For model analysis, we report the overall performance at the box level, point level, and the average. SPT$_{\text{LoRA}}$ based on ViT-B is leveraged here by default.

\begin{table}[!h]
\belowrulesep=0pt
\aboverulesep=0pt
\renewcommand{\arraystretch}{1.1}
\centering
\small
\setlength{\tabcolsep}{0.7mm}
\caption{Ablation study of each component in the proposed SPT (\%). VRA is an abbreviation for VRA-Adapter.}\label{tab:ablation_vit_b}
\begin{tabular}{cc|cc|cc|cc}
\toprule
\multirow{2}{*}{SDT} & \multirow{2}{*}{VRA}& \multicolumn{2}{c|}{box-level } & \multicolumn{2}{c|}{point-level} & \multicolumn{2}{c}{Avg.}\\
\cmidrule{3-8} 
& & mIoU & mBIoU & mIoU & mBIoU & mIoU & mBIoU \\
\midrule
 \multicolumn{2}{c|}{zero-shot} & 59.7 & 54.3 & 44.3 & 39.8 & 52.0 & 47.0 \\
 \multicolumn{2}{c|}{PEFT} & 67.5 & 61.8 & 65.6 & 60.2 & 66.5 & 61.0 \\
 \midrule
 \checkmark & & 68.6 & 62.9 & 66.7 & 61.6 & 67.6 & 62.2 \\
& \checkmark & 68.1 & 62.3 & 65.3 & 60.1 & 66.7 & 61.2 \\
\checkmark & \checkmark & \textbf{69.4} & \textbf{63.5} & \textbf{67.4} & \textbf{62.2} & \textbf{68.4} & \textbf{62.8} \\
\bottomrule
\end{tabular}
\end{table}

\paragraph{Ablation study.}
To verify the effectiveness of each component in our SPT, we introduce zero-shot SAM and SAM tuned by PEFT method as competitors. As shown in Table \ref{tab:ablation_vit_b}, we can see that each component can substantially contribute to the performance gains. Benefited from the nature of draft-then-refine process, the proposed self-draft tuning (SDT) can outperform conventional PEFT method with over 1.0\% improvement for both metrics, indicating the advantage of SDT. Combined with VRA-Adapter, the gains can reach to 1.9\% mIoU and 1.8\% mBIoU, demonstrating the positive effect of VRA-Adapter. We also present the cost and efficiency analysis for each component. From Table \ref{tab:ablation_efficient_vit_b}, we can observe that the proposed SDT and VRA-Adapter only introduce small extra learnable parameters, leading to marginal memory assumption increase during training and tiny inference throughput decrease. These evidences can well demonstrate the effectiveness of each component in SPT.

\begin{table}[!t]
\belowrulesep=0pt
\aboverulesep=0pt
\renewcommand{\arraystretch}{1.1}
\centering
\small
\setlength{\tabcolsep}{0.7mm}

\caption{Cost and efficiency analysis of each component.}\label{tab:ablation_efficient_vit_b}
\begin{tabular}{cc|c|c|c|c|c}
\toprule
\multirow{2}{*}{SDT} & \multirow{2}{*}{VRA} & All & Trainable &ratio &memory  &throughput \\
& & (M) & (M) & (\%) & (M) & (fps) \\
\midrule
 \multicolumn{2}{c|}{zero-shot} &89.393 &- &- &- & 8.67\\
 \multicolumn{2}{c|}{PEFT}   &89.719 &0.326 &0.364 & 7904 & 8.69\\
 \midrule
 \checkmark & &93.635 &0.371 &0.396 &7924 & 8.27\\
 & \checkmark  &89.725 &0.327 &0.364 &7907 & 8.46\\
\checkmark & \checkmark &93.636 &0.372 &0.397 &7926 & 7.98\\

\bottomrule
\end{tabular}
\end{table}

\begin{table}[!t]
\belowrulesep=0pt
\aboverulesep=0pt
\renewcommand{\arraystretch}{1.1}
\centering
\small
\setlength{\tabcolsep}{0.7mm}
\caption{The impact of the self-draft tuning (\%).}\label{tab:ablation_draft_vit_b}
\begin{tabular}{c|cc|cc|cc}
\toprule
& \multicolumn{2}{c|}{box-level} & \multicolumn{2}{c|}{point-level} & \multicolumn{2}{c}{Avg.} \\

\cmidrule{2-7}
\multirow{-2}*{Method}  & mIoU & mBIoU & mIoU & mBIoU & mIoU & mBIoU \\
\midrule
 zero-shot SAM & 59.7 & 54.3 & 44.3 & 39.8 & 52.0 & 47.0 \\
 zero-shot SDT & 59.9 & 55.1 & 46.0 & 41.7 & 53.0 & 48.4 \\
 \midrule
 SPT w $\mathbf{D}_{\text{mask}}\ 2\times$ & 67.8 & 62.0 & 66.2 & 61.2 & 67.0 & 61.6 \\
 SPT w $\mathbf{D}_{\text{draft}}\times1$& \text{69.4} & \text{63.5} & \text{67.4} & \text{62.2} & \textbf{68.4} & \text{62.8} \\
 SPT w $\mathbf{D}_{\text{draft}}\times2$& \text{68.8} & \text{63.2} & \text{67.5} & \text{62.5} & \text{68.1} & \textbf{62.9} \\
 
\bottomrule
\end{tabular}
\end{table}

\paragraph{Analysis of the self-draft tuning.} First, we further explore the benefits of the draft-then-refine process by comparing zero-shot SAM with zero-shot SDT, which applies SDT to SAM without PEFT tuning. As shown in Table \ref{tab:ablation_draft_vit_b}, zero-shot SDT substantially achieves improvements across all metrics, resulting in gains of 1.0\% in mIoU and 1.4\% in mBIoU on average. The results from this zero-shot comparison effectively highlight the superiority of SDT, as confirmed by the ablation study. Additionally, we analyze the impact of using more $\mathbf{D}_{\text{draft}}$ with our SPT, as well as the effect of employing a single $\mathbf{D}_{\text{mask}}$ that is applied twice during both training and inference with shared LoRA parameters (denoted by ``SPT w $\mathbf{D}_{\text{mask}} \ 2\times$''). We can see that the performance of ``SPT w $\mathbf{D}_{\text{draft}}\times2$'' is comparable to that of ``SPT w $\mathbf{D}_{\text{draft}}\times1$'', indicating that a single $\mathbf{D}_{\text{draft}}$ is sufficient to achieve remarkable performance. However, the performance of ``SPT w $\mathbf{D}_{\text{mask}} \ 2\times$'' is  inferior to that of ``SPT w $\mathbf{D}_{\text{draft}}\times1$'', highlighting the necessity of the draft decoder in our proposed strategy.

\begin{table}[!ht]
\belowrulesep=0pt
\aboverulesep=0pt
\renewcommand{\arraystretch}{1.1}
\centering
\small
\setlength{\tabcolsep}{0.8mm}
\caption{The impact of accommodating PEFT (\%).}\label{tab:ablation_peft_vit_b}
\begin{tabular}{c|cc|cc|cc}
\toprule
& \multicolumn{2}{c|}{box-level} & \multicolumn{2}{c|}{point-level} & \multicolumn{2}{c}{Avg.} \\

\cmidrule{2-7}
 \multirow{-2}{*}{PEFT} & mIoU & mBIoU & mIoU & mBIoU & mIoU & mBIoU \\
\midrule

 None & 59.9 & 55.1 & 46.0 & 41.7 & 53.0 & 48.4 \\
only $\mathbf{D}_{\text{refine}}$& 63.0 & 58.3 & 58.6 & 54.5 & 60.8 & 56.4 \\
only $\mathbf{D}_{\text{draft}}$& 61.6 & 57.0 & 62.7 & 57.1 & 62.1 & 57.0 \\
 no decoder & 66.5 & 60.5 & 61.6 & 55.9 & 64.0 & 58.2 \\
 no encoder & 63.5 & 59.3 & 61.4 & 57.6 & 62.4 & 58.4 \\
 All & \textbf{68.6} & \textbf{62.9} & \textbf{66.7} & \textbf{61.6} & \textbf{67.6} & \textbf{62.2} \\
\bottomrule
\end{tabular}
\end{table}

\begin{table}[!ht]
\belowrulesep=0pt
\aboverulesep=0pt
\renewcommand{\arraystretch}{1.1}
\centering
\small
\setlength{\tabcolsep}{0.7mm}
\caption{The impact of VRA-Adapter (\%).}\label{tab:ablation_vatt_vit_b}
\begin{tabular}{c|cc|cc|cc}
\toprule
& \multicolumn{2}{c|}{box-level} & \multicolumn{2}{c|}{point-level} & \multicolumn{2}{c}{Avg.} \\
\cmidrule{2-7}
 \multirow{-2}{*}{VRA-Adapter} & mIoU & mBIoU & mIoU & mBIoU & mIoU & mBIoU \\
 \midrule
 None & 68.6 & 62.9 & 66.7 & 61.6 & 67.6 & 62.2\\
$\mathbf{TW}_1$ & 68.8 & 63.1 & 67.5 & \textbf{62.4} & 68.2 & \textbf{62.8} \\
 $\mathbf{TW}_2$ & 69.0 & 63.2 & 67.0 & 61.9 & 68.0 & 62.5 \\
$\mathbf{TW}_1$ + $\mathbf{TW}_2$ & \textbf{69.4} & \textbf{63.5} & \textbf{67.4} & 62.2 & \textbf{68.4} & \textbf{62.8} \\
\bottomrule
\end{tabular}
\end{table}

\begin{table}[!ht]
\belowrulesep=0pt
\aboverulesep=0pt
\renewcommand{\arraystretch}{1.1}
\centering
\small
\caption{Analysis of automatic prompts (\%).}\label{tab:saa_vit_b}
\begin{tabular}{c|ccc}
\toprule
Method & AUROC & AP & max-F1\\
\midrule
zero-shot & 73.6 & 31.1 & 41.2 \\
Ours & \textbf{74.8} & \textbf{33.1} & \textbf{42.5} \\
\bottomrule
\end{tabular}
\end{table}

\begin{figure*}[ht]
\centering
\small
\includegraphics[width=.75\linewidth]{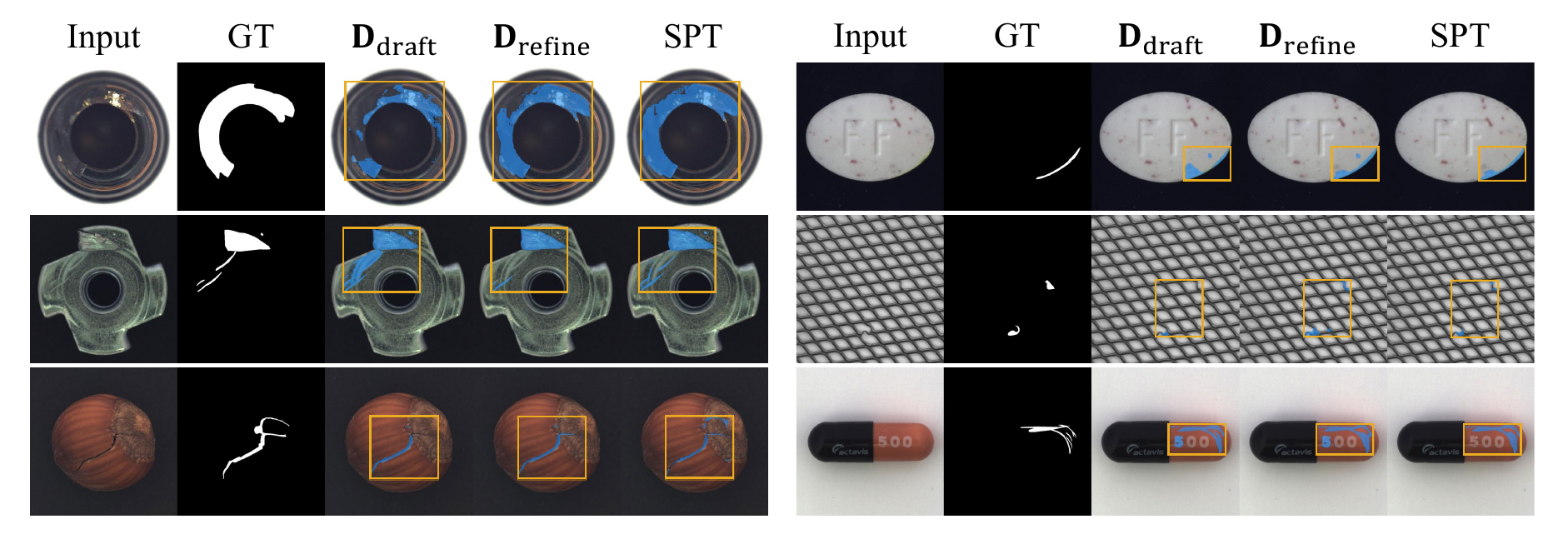}
\caption{Examples for comparison among components in SPT. We use $\mathbf{D}_{\text{draft}}$ and $\mathbf{D}_{\text{refine}}$ for analysing SDT and SPT for analysing VRA-Adapter. GT denotes the ground truth.}
\label{fig5: vis_draft}
\end{figure*}

\begin{figure*}[ht]
\centering
\small
\includegraphics[width=.75\linewidth]{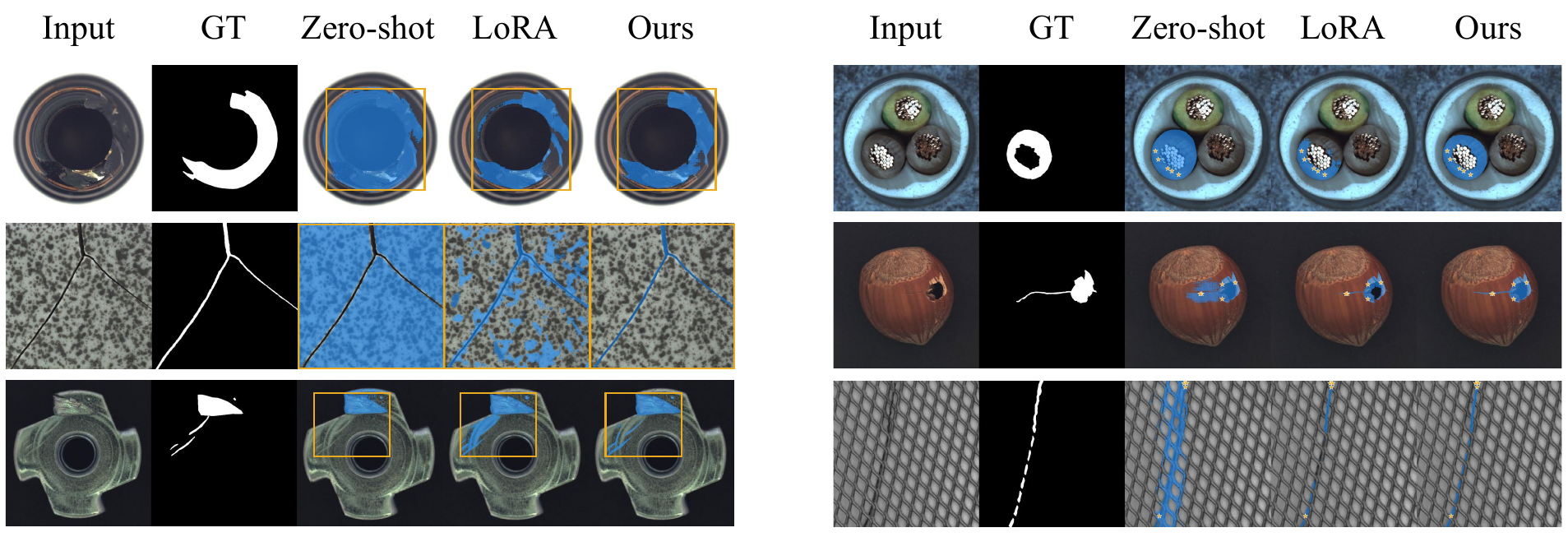}
\caption{Qualitative analysis using different prompts. We provide examples with box-level prompts (left) and point-level prompts (right). GT means ground truth.}
\label{fig7:vis_box_point}
\vspace{-5pt}
\end{figure*}

\paragraph{Accurate draft enhances the performance.} To gain further insights into the draft-then-refine process, we investigate how the perception degradation of each part of the model affects the overall segmentation performance. As shown in Table \ref{tab:ablation_peft_vit_b}, we leave out PEFT modules, \textit{i.e.,} LoRA here, in the image encoder $\mathbf{E}$, draft decoder $\mathbf{D}_{\text{draft}}$, and mask decoder $\mathbf{D}_{\text{refine}}$ separately. We can observe that (1) compared with ``only $\mathbf{D}_{\text{refine}}$'', ``only $\mathbf{D}_{\text{draft}}$'' shows better performance. Compared with ``no PEFT'' (\textit{i.e.,} None), ``only $\mathbf{D}_{\text{draft}}$'' exhibits significant performance gains. These evidences demonstrate that the performance can be enhanced with more accurate draft result. (2) compared with ``no decoder'', ``no encoder'' suffers from more performance drops, indicating the importance of adapting representation of SAM to the industrial images. It also implicitly reveal that draft on inaccurate information can still encounter segmentation degradation. (3) enhancing the perception ability for all parts can achieve the optimal adaptation performance. 

\paragraph{Analysis of the VRA-Adapter.} We investigate the impact of VRA-Adapter with respect to its replacement. We consider three cases that placing it in $\mathbf{TW}_1$, $\mathbf{TW}_2$, or both. As illustrated in Table \ref{tab:ablation_vatt_vit_b}, by comparing with no VRA-Adapter, we observe that the performance can be enhanced as long as the VRA-Adapter is applied. Besides, we can see that (1) placing it in an earlier layer is more beneficial than in a later layer, and (2) using it in both layers yields the best performance. These observations effectively demonstrate the positive impact of incorporating visual relations into the decoder.

\paragraph{SPT also works using automatic prompts.} Here we evaluate the effectiveness of our method when using automatic prompts generated by GroundingDINO as SAA+ \cite{saa}. Specifically, we directly replace the SAM model in SAA+ with our constructed model using the official code\footnote{https://github.com/caoyunkang/Segment-Any-Anomaly}.
For convenience and direct comparison, we employ the same human-crafted prompts as those used in SAA+, even though these prompts are known to be sensitive to different models and datasets. As illustrated in Table \ref{tab:saa_vit_b}, our method can consistently outperform the zero-shot SAM across different metrics with a maximal performance improvement of 2.0\% (AP). These results well demonstrate the effectiveness of our method using the automatic prompts.

\subsection{Qualitative Analysis}
\paragraph{Visualization analysis for components in SPT.} For intuitively understanding the advantage of each component in SPT, we visualize the anomaly mask generated by $\mathbf{D}_{\text{draft}}$, $\mathbf{D}_{\text{refine}}$ and SPT for analysing SDT and VRA-Adapter. As shown in Fig. \ref{fig5: vis_draft}, the masks produced by $\mathbf{D}_{\text{draft}}$ provide a coarse outline of the anomalous regions, but they may include some extraneous areas or miss certain parts. After going through the refine step, the anomaly mask generated by $\mathbf{D}_{\text{refine}}$ becomes more accurate and precise. Enhanced by VRA-Adapter, our SPT can capture more accurate mask with high correlation, compared with $\mathbf{D}_{\text{refine}}$. These evidences well support the effectiveness of each component.

\paragraph{Visualization of anomaly segmentation using different prompts.} To demonstrate the superiority of our method in the generalization using different prompts, we visualize and compare segmentation results of different methods. As shown in Fig. \ref{fig7:vis_box_point}, the original SAM model lacks the ability to recognize defects, often resulting in localizing continuous regions that are not anomalies. Although LoRA improves the performance to some extent, it performs worse than our method. For example, in the point-level prompt mode, LoRA tends to simply segment areas exactly covered by points. In contrast, our method can discover anomalies that the points do not fully covers, demonstrating its superior robustness in such scenarios. Such generalization can be largely attributed to the designed visual-relation-aware adapter which can enhance the relationship among anomalous regions, leading to consistent and accurate anomaly segmentation.

\section{Conclusion}
In this paper, we propose a novel Self-Perception Tuning (SPT) method to adapt SAM for practical anomaly segmentation scenarios. Unlike the original SAM and conventional parameter-efficient fine-tuning (PEFT) methods, our approach aims to enhance SAM's perception capabilities during adaptation to address domain shift issues. Specifically, SPT incorporates a self-drafting tuning strategy, which first generates a coarse draft of the anomaly mask and then undergoes a refinement process. Additionally, we introduce a visual-relation-aware adapter to improve the perception of discriminative relation information for mask generation. Extensive experimental results on several benchmark datasets show that our SPT method achieves state-of-the-art performance, validating its effectiveness.

\appendix

\setcounter{secnumdepth}{0}
\section{Acknowledgments}
This work was supported by National Science and Technology Major (No. 2022ZD0119401), National Natural Science Foundation of China (No. 61925107, 62271281, 62021002).

\bibliography{aaai25}

\clearpage

\setcounter{secnumdepth}{2}


\section{Datasets}

\subsection{Dataset for Fine-Tuning}
The training dataset is collected from real production environments. The categories of anomaly are diverse, including 3C (Computer, Communication, and Consumer Electronics), photovoltaics, graphite film, display screens, printing, and lithium batteries. 
All of these samples contain defects and are annotated with pixel-level labels. We utilize this dataset to fine-tune the SAM model, which is subsequently tested on six benchmark datasets.

\subsection{Benchmark Datasets}

\paragraph{Data Preprocessing.}
To prepare the data for the different evaluation modes in our anomaly segmentation models, we applied the following preprocessing steps.
For the ``one box'' mode, the preprocessing involves generating the minimum bounding box that covers all defects in an image, which then serves as the prompt.
For the ``multiple boxes'' mode, we segment the non-connected defect masks into distinct regions, each representing a separate defect. This process involved identifying connected regions in the mask and discarding areas less than 50 pixels.
For the ``points'' mode, we randomly sampled either 5 or 10 points within the defective areas to serve as prompts, which shares the same examples as ``multiple boxes''.



\paragraph{MVTec-AD.} MVTec-AD \cite{mvtec} is a comprehensive dataset for unsupervised anomaly detection in industrial settings, containing 5,354 high-resolution images of 15 different object and texture categories in the test set. After data preprocessing, there are 1,258 examples for the ``one box'' evaluation and 1,884 examples for the ``multiple boxes'' and ``point'' evaluations modes. 


\paragraph{VisA.} The Visual Industrial Surface Anomaly (VisA) dataset \cite{visa} consists of 12 subsets. It contains various anomalies such as surface defects and structural flaws during testing. After data preprocessing, the test set consists of 1,200 examples for the ``one box'' evaluation mode and 2,456 examples in the ``multiple boxes'' mode and the ``point'' evaluation modes.

\paragraph{MTD.} The MTD (Magnetic Tile Defects) dataset \cite{mtd} focuses on surface defects in magnetic tiles. After data preprocessing, we have 387 examples for the ``one box'' evaluation mode and 423 examples for the ``multiple boxes'' mode and ``point'' evaluation modes.

\paragraph{KSDD2.} The Kolektor Surface Defect Dataset 2 (KSDD2) \cite{ksdd2} provides various types of surface defects in an industrial setting. After data preprocessing, the test set consists of 110 images in the ``one box'' evaluation mode and 119 images in the ``multiple boxes'' and ``point'' evaluation modes.


\paragraph{BTAD.} The BeanTech Anomaly Detection (BTAD) dataset \cite{btad} focuses on surface anomalies in manufactured parts. After data preprocessing, it has 279 images for the ``one box'' evaluation mode and 692 images for the ``multiple boxes'' and ``point'' evaluation modes.

\paragraph{MPDD.} The Metal Parts Defect Detection (MPDD) dataset \cite{mpdd} consists of 1346 images of metal parts. After data preprocessing, there are 282 images in the ``one box'' evaluation mode and 471 images in the ``multiple boxes'' and ``point'' evaluation modes.

\section{Additional Introduction to PEFT Methods}

\paragraph{LoRA}
Low-Rank Adaptation (LoRA) \cite{lora} is one of the most popular PEFT methods to tune large foundation models. It reduces the number of trainable parameters by decomposing the weight matrices of the model into low-rank components. The key idea is to constrain the update of the pre-trained weight matrix$\mathbf{W}_0 \in \mathbb{R}^{d \times k}$ by representing the latter with a low-rank decomposition $\Delta\mathbf{W} \in \mathbb{R}^{d \times k} = \mathbf{B}\mathbf{A}$ where $\mathbf{B} \in \mathbb{R}^{d \times r}$, $\mathbf{A} \in \mathbb{R}^{r \times k}$, and the rank $r \ll \min(d, k)$. Given the input $\mathbf{x}$, it yields:
\begin{equation}
\label{eq:lora}
\mathbf{W}\mathbf{x} = \mathbf{W}_0\mathbf{x} + \Delta\mathbf{W}\mathbf{x} = \mathbf{W_0}\mathbf{x} + \mathbf{B}\mathbf{A}\mathbf{x}
\end{equation}

\paragraph{DoRA}
Weight-Decomposed Low-Rank Adaptation (DoRA) \cite{dora} extends LoRA by decomposing the pre-trained weights into magnitude and direction components, enhancing the learning capacity and training stability:
\begin{equation}
\label{eq:dora}
\mathbf{W}\mathbf{x} = \underline{\mathbf{m}} \frac{\mathbf{V}+\Delta \mathbf{V}}{\|\mathbf{V}+\Delta \mathbf{V}\|_c}\mathbf{x} = \underline{\mathbf{m}}  \frac{\mathbf{W}_0+\underline{\mathbf{BA}}}{\|\mathbf{W}_0+\underline{\mathbf{BA}}\|_c}\mathbf{x}
\end{equation}
where the underlined parameters denote trainable. $\mathbf{B}\in \mathbb{R}^{d \times r}$ and $\mathbf{A} \in \mathbb{R}^{r \times k}$ are initialized similarly as LoRA. $\mathbf{m}$ is the magnitude vector, and $\mathbf{V}$ is the direction matrix. They satisfy $\mathbf{W}_0 = \mathbf{m} \frac{\mathbf{V}}{\|\mathbf{V}\|_c}$.

\paragraph{Adapter}
Adapter modules \cite{adapter} introduce small neural networks that are inserted into each layer of the pre-trained model. These modules enable task-specific fine-tuning while keeping the original model parameters frozen. Given an input $\mathbf{x} \in \mathbb{R}^d$ to a layer, an adapter module consists of a down-projection $\mathbf{W}_{\text{down}} \in \mathbb{R}^{d \times r}$ and an up-projection $\mathbf{W}_{\text{up}} \in \mathbb{R}^{r \times d}$ where $r \ll d$. The output $\mathbf{o}$ of the adapter module is:
\begin{equation}
\label{eq:adapter}
\mathbf{o} = \mathbf{x} + \mathbf{W}_{\text{up}} \sigma(\mathbf{W}_{\text{down}}\mathbf{x})
\end{equation}
where $\sigma$ is a non-linear activation function.

\begin{table*}[!ht]
\renewcommand{\arraystretch}{1.2}
\belowrulesep=0pt
\aboverulesep=0pt
\centering
\small
\caption{Performance comparison of different fine-tuning methods on the anomaly segmentation task. Results are reported as mIoU / mBIoU for each prompt type.}
\label{tab:decoder_finetuning}
\begin{tabular}{c|cc|cc|cc|cc|cc}
\toprule
& \multicolumn{2}{c|}{One box} & \multicolumn{2}{c|}{Multiple boxes} & \multicolumn{2}{c|}{Point=5} & \multicolumn{2}{c|}{Point=10} & \multicolumn{2}{c}{Avg.}\\
\cmidrule{2-11}
 \multirow{-2}*{Method} & mIoU & mBIoU & mIoU & mBIoU & mIoU & mBIoU & mIoU & mBIoU & mIoU & mBIoU \\
\midrule
decoder-full & 39.6 & 37.0 & 41.6 & 39.5 & 41.6 & 38.4 & 47.0 & 42.4 & 42.5 & 39.3 \\
decoder-LoRA & 58.9 & 54.1 & 65.6 & 61.5 & 52.8 & 49.6 & 66.7 & 61.1 & 61.0 & 56.6 \\
\midrule
\rowcolor{light-gray} Ours & 67.1 & 60.7 & 71.7 & 66.3 & 70.1 & 64.5 & 64.6 & 59.8 & 68.4 & 62.8 \\
\bottomrule
\end{tabular}
\end{table*}

\section{Additional Experimental Results}
\subsection{Analysis of Decoder-Only Fine-Tuning}

We conduct experiments to explore the performance of fine-tuning only the transformer decoder, comparing two setups: (1) fully fine-tuning the decoder (decoder-full), and (2) applying LoRA to the decoder (decoder-LoRA). Table~\ref{tab:decoder_finetuning} summarizes the results across different prompt types.
The results show that decoder-full performs significantly worse than both decoder-LoRA and our proposed method. We believe this is due to two main factors: (1) the large domain gap between SA-1B, the dataset used to train SAM, and the industrial anomaly images in our task, and (2) the limited number of anomaly images available for training in our setting. For example, LISA~\cite{lisa} achieves better results with decoder fine-tuning due to its smaller domain gap (natural images) and the use of larger datasets like ADE20K, COCO-Stuff, and Llava-Instruct-150k.
In contrast, for industrial anomaly segmentation, fully fine-tuning the decoder risks overfitting to the small training dataset, reducing generalization to unseen datasets. This highlights the importance of our approach, which effectively balances parameter-efficient fine-tuning and robustness across diverse scenarios.

\begin{figure}[!t]
    \centering
    \includegraphics[width=\linewidth]{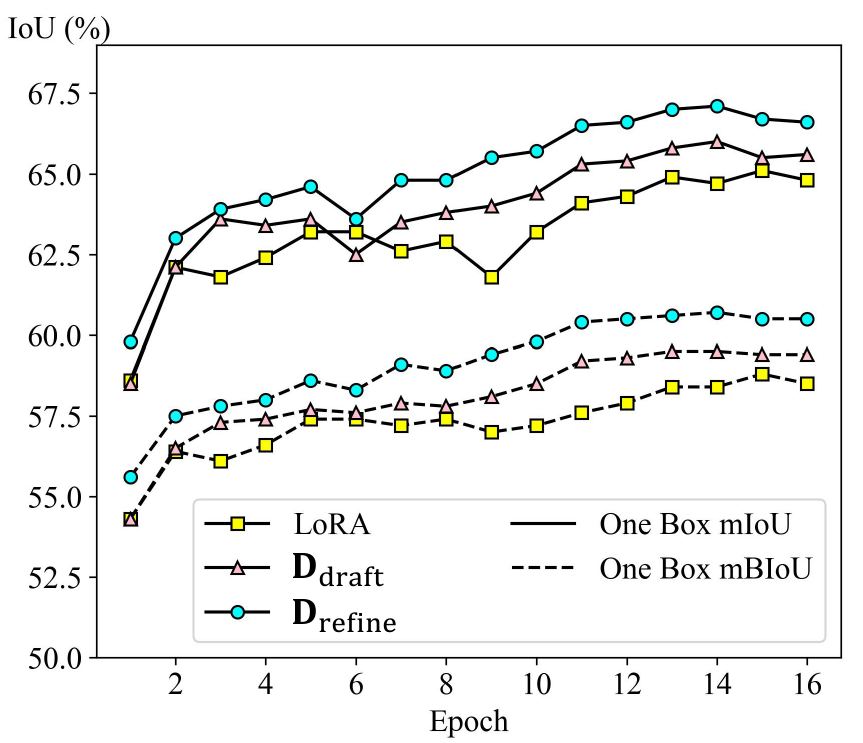}
    \caption{Learning curves during training.}
    \label{fig:learning_curve}
\end{figure}

\begin{figure}[!t]
    \centering
    \includegraphics[width=\linewidth]{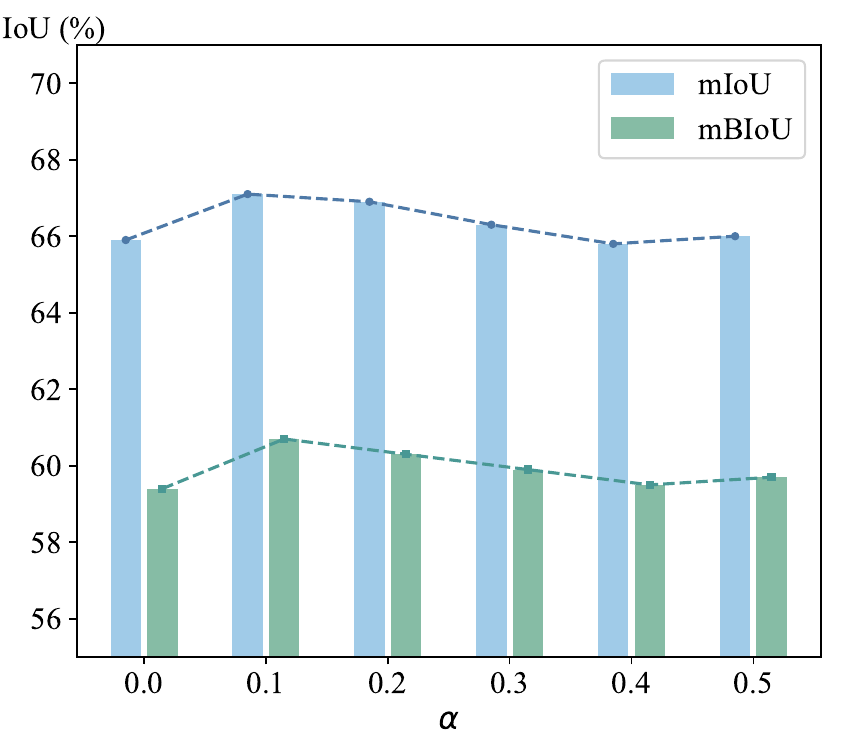}
    \caption{Sensitive analysis of the hyperparameter $\alpha$.}
    \label{fig:Sensitive_alpha}
\end{figure}

\subsection{Learning curves during training.}
 During training, we use metrics from the evaluation mode of one box to represent the learning curves. In Figure~\ref{fig:learning_curve}, we analyze the performance trends of LoRA, the output of $\mathbf{D}_{\text{draft}}$, and the final refined output of $\mathbf{D}_{\text{refine}}$. We observe that $\mathbf{D}_{\text{draft}}$ performs better than LoRA, indicating that enhancing self-perception can improve the original SAM process. Additionally, both $\mathbf{D}_{\text{draft}}$ and $\mathbf{D}_{\text{refine}}$ consistently outperform LoRA, demonstrating the superiority of our method.

\subsection{Hyperparameter analysis.}
We provide results of $\alpha$ in VRA-Adapter with values ranging from 0.0 to 0.5. Results in Fig. \ref{fig:Sensitive_alpha} well demonstrate the robustness of $\alpha$.

\begin{table*}[!ht]
\belowrulesep=0pt
\aboverulesep=0pt
\renewcommand{\arraystretch}{1.2}
\centering
\small
\caption{Performance comparison with ViT-B backbone under different evaluation modes (\%). “Avg.” denotes the average scores of four kinds of prompts. Values in blue denotes the improvement compared with the zero-shot SAM.}\label{tab:vit_b_appendix}
\begin{tabular}{c|cc|cc|cc|cc|cc}
\toprule
  & \multicolumn{2}{c|}{One box} & \multicolumn{2}{c|}{Multiple boxes} & \multicolumn{2}{c|}{Point=5} & \multicolumn{2}{c|}{Point=10} & \multicolumn{2}{c}{Avg.}\\
\cmidrule{2-11}
 \multirow{-2}*{Method} & mIoU & mBIoU & mIoU & mBIoU & mIoU & mBIoU & mIoU & mBIoU & mIoU & mBIoU \\
\midrule
zero-shot & 56.3 & 50.7 & 63.0 & 57.9 & 44.1 & 39.8 & 44.4 & 39.7 & 52.0 & 47.0 \\
\midrule
SAMAdapter & 60.2 & 53.4 & 67.3 & 61.4 & 56.1 & 50.5 & 59.2 & 52.9 & 60.7 & 54.6 \\
AdaptFormer & 64.8 & 58.9 & 69.2 & 64.4 & 62.6 & 57.4 & 65.9 & 60.3 & 65.6 & 60.3 \\
NOLA & 63.8 & 57.7 & 69.4 & 64.2 & 61.9 & 56.6 & 66.5 & 60.5 & 65.4 & 59.8 \\
BitFit & 61.4 & 56.1 & 65.9 & 61.6 & 60.6 & 56.1 & 67.2 & 61.6 & 63.8 & 58.9 \\
LoRA & 65.1 & 58.8 & 69.9 & 64.8 & 62.4 & 57.5 & 68.7 & 62.9 & 66.5 & 61.0 \\
DoRA & 65.3 & 58.8 & 70.1 & 64.7 & 62.8 & 57.9 & 67.7 & 62.0 & 66.5 & 60.9 \\
Adapter & 65.1 & 59.2 & 70.0 & 65.3 & 61.8 & 56.9 & 67.2 & 61.4 & 66.0 & 60.7\\
\midrule
\rowcolor{light-gray} Ours & \textbf{66.9} & \textbf{60.6} & \textbf{71.4} & \textbf{66.4} & \textbf{65.3} & \textbf{60.4} & \textbf{70.5} & \textbf{64.9} & \textbf{68.5} & \textbf{63.1} \\

\bottomrule
\end{tabular}
\end{table*}

\begin{table*}[!ht]
\belowrulesep=0pt
\aboverulesep=0pt
\renewcommand{\arraystretch}{1.2}
\centering
\small
\caption{Performance comparison with ViT-L backbone under different evaluation modes (\%). “Avg.” denotes the average scores of four kinds of prompts. Values in blue denotes the improvement compared with the zero-shot SAM.}\label{tab:vit_l}
\begin{tabular}{c|cc|cc|cc|cc|cc}
\toprule
  & \multicolumn{2}{c|}{One box} & \multicolumn{2}{c|}{Multiple boxes} & \multicolumn{2}{c|}{Point=5} & \multicolumn{2}{c|}{Point=10} & \multicolumn{2}{c}{Avg.}\\
\cmidrule{2-11}
 \multirow{-2}*{Method} & mIoU & mBIoU & mIoU & mBIoU & mIoU & mBIoU & mIoU & mBIoU & mIoU & mBIoU \\
 \midrule
zero-shot & 58.7 & 52.4 & 65.0 & 59.5 & 41.6 & 38.1 & 39.8 & 36.1 & 51.3 & 46.5\\
\midrule
SAMAdapter & 63.8 & 57.5 & 68.2 & 63.0 & 62.8 & 57.2 & 66.9 & 60.8 & 65.4 & 59.6 \\
AdaptFormer & 67.5 & 61.6 & 71.4 & 66.5 & 63.5 & 58.2 & 67.7 & 62.2 & 67.5 & 62.1 \\
NOLA & 68.4 & 61.8 & 72.5 & 67.1 & 63.8 & 58.6 & 68.5 & 63.0 & 68.3 & 62.6\\
BitFit & 67.4 & 61.1 & 70.8 & 65.8 & 63.1 & 57.9 & 69.2 & 63.6 & 67.6 & 62.1 \\
LoRA & 68.1 & 62.1 & 71.8 & 67.1 & 64.8 & 59.7 & 70.0 & 64.4 & 68.7 & 63.3 \\
DoRA & 68.3 & 61.9 & 72.5 & 67.3 & 63.9 & 58.7 & 68.7 & 63.2 & 68.4 & 62.8 \\
Adapter & 68.2 & 61.8 & 71.9 & 66.8 & 63.5 & 58.2 & 68.4 & 62.9 & 68.0 & 62.4 \\
\midrule
\rowcolor{light-gray} Ours & \textbf{69.1} & \textbf{63.0} & \textbf{73.3} & \textbf{68.3} & \textbf{66.1} & \textbf{60.9} & \textbf{70.3} & \textbf{65.0} & \textbf{69.7} & \textbf{64.3} \\
\bottomrule
\end{tabular}
\end{table*}

\begin{table*}[!ht]
\belowrulesep=0pt
\aboverulesep=0pt
\renewcommand{\arraystretch}{1.2}
\centering
\small
\caption{Performance comparison with ViT-H backbone under different evaluation modes (\%). “Avg.” denotes the average scores of four kinds of prompts. Values in blue denotes the improvement compared with the zero-shot SAM.}\label{tab:vit_h}
\begin{tabular}{c|cc|cc|cc|cc|cc}
\toprule
& \multicolumn{2}{c|}{One box} & \multicolumn{2}{c|}{Multiple boxes} & \multicolumn{2}{c|}{Point=5} & \multicolumn{2}{c|}{Point=10} & \multicolumn{2}{c}{Avg.}\\

\cmidrule{2-11}
 \multirow{-2}*{Method} & mIoU & mBIoU & mIoU & mBIoU & mIoU & mBIoU & mIoU & mBIoU & mIoU & mBIoU \\
\midrule
zero-shot & 57.6 & 51.3 & 64.2 & 58.7 & 46.8 & 42.6 & 50.8 & 45.8 & 54.9 & 49.6\\
\midrule
SAMAdapter & 63.5 & 57.1 & 68.2 & 63.0 & 60.5 & 55.0 & 65.0 & 59.0 & 64.3 & 58.5 \\
AdaptFormer & 67.3 & 61.1 & 71.7 & 66.6 & 63.1 & 58.2 & 68.6 & 63.2 & 67.7 & 62.3 \\
BitFit & 64.4 & 58.4 & 67.8 & 62.9 & 60.9 & 56.0 & 67.4 & 61.5 & 65.1 & 59.7 \\
LoRA & 68.1 & 61.7 & 71.8 & 66.6 & 65.0 & 59.7 & 69.8 & 64.1 & 68.7 & 63.0 \\
DoRA & 67.8 & 61.8 & 71.5 & 66.6 & 64.0 & 59.0 & 69.8 & 64.3 & 68.3 & 62.9 \\
Adapter & 66.8 & 60.4 & 71.1 & 66.0 & 63.2 & 58.0 & 67.7 & 62.0 & 67.2 & 61.6 \\
\midrule
\rowcolor{light-gray} Ours & \textbf{69.7} & \textbf{63.2} & \textbf{73.7} & \textbf{68.5} & \textbf{66.5} & \textbf{61.3} & \textbf{72.2} & \textbf{66.6} & \textbf{70.5} & \textbf{64.9} \\

\bottomrule
\end{tabular}
\end{table*}

\subsection{Comparison with Baseline Methods}
We include more performance comparison results with additional baseline methods and also based on the ViT-L and ViT-H backbones. Alongside the baseline methods presented in the main paper, we also provide results for additional parameter-efficient fine-tuning (PEFT) methods, including SAMAdapter \cite{samadapter}, AdapterFormer \cite{adaptformer}, NOLA \cite{nola}, and BitFit \cite{bitfit}. Due to GPU limitations, we are unable to implement NOLA on the ViT-H backbone. Table \ref{tab:vit_b_appendix}, Table \ref{tab:vit_l} and Table \ref{tab:vit_h} summarize the performance of these methods and our SPT, offering a comprehensive view of SPT's effectiveness. It is obvious that our method exhibits significant improvements over various baseline approaches. 

\begin{figure*}[ht!]
\centering
\includegraphics[width=0.93\textwidth]{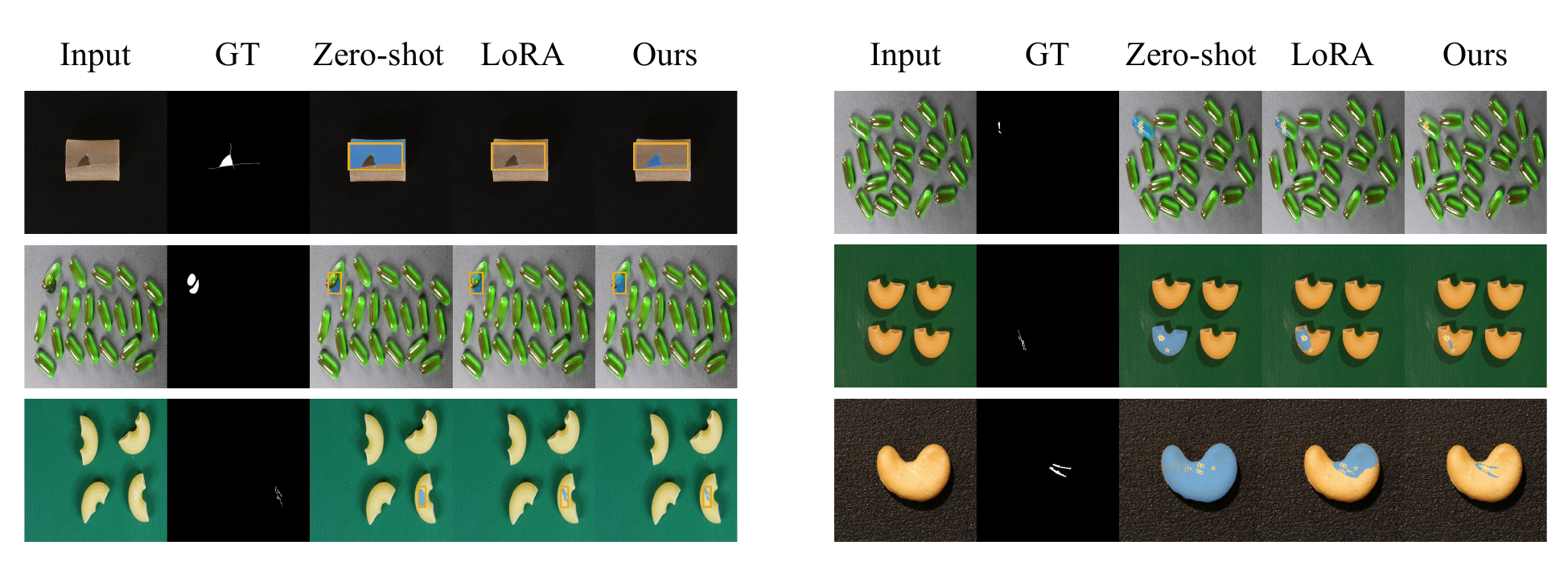}
        \caption{Qualitative analysis on VisA using different prompts. We provide examples with box-level prompts (left) and point-level prompts (right). GT means ground truth.}
        \label{fig:vis_visa}
\end{figure*}

\begin{figure*}[ht!]
\centering
\includegraphics[width=0.93\textwidth]{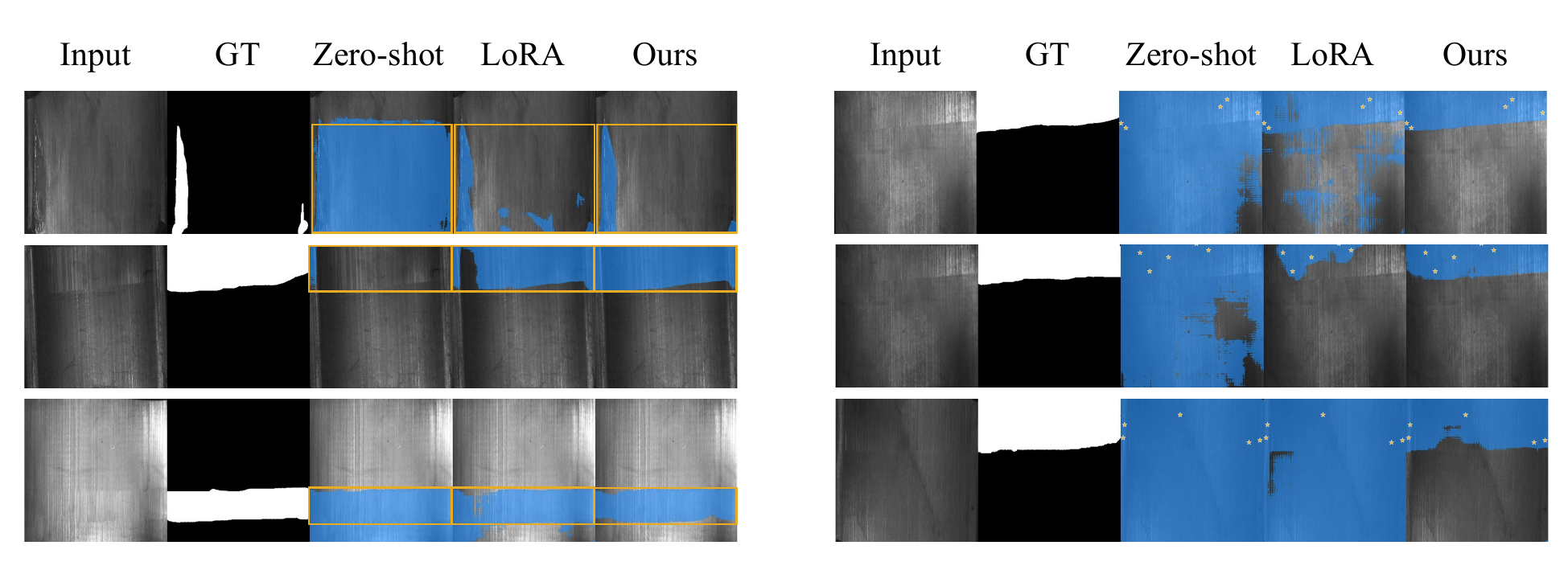}
        \caption{Qualitative analysis on MTD using different prompts. We provide examples with box-level prompts (left) and point-level prompts (right). GT means ground truth.}
        \label{fig:vis_mtd}
\end{figure*}

\begin{figure*}[ht!]
\centering
\includegraphics[width=0.93\textwidth]{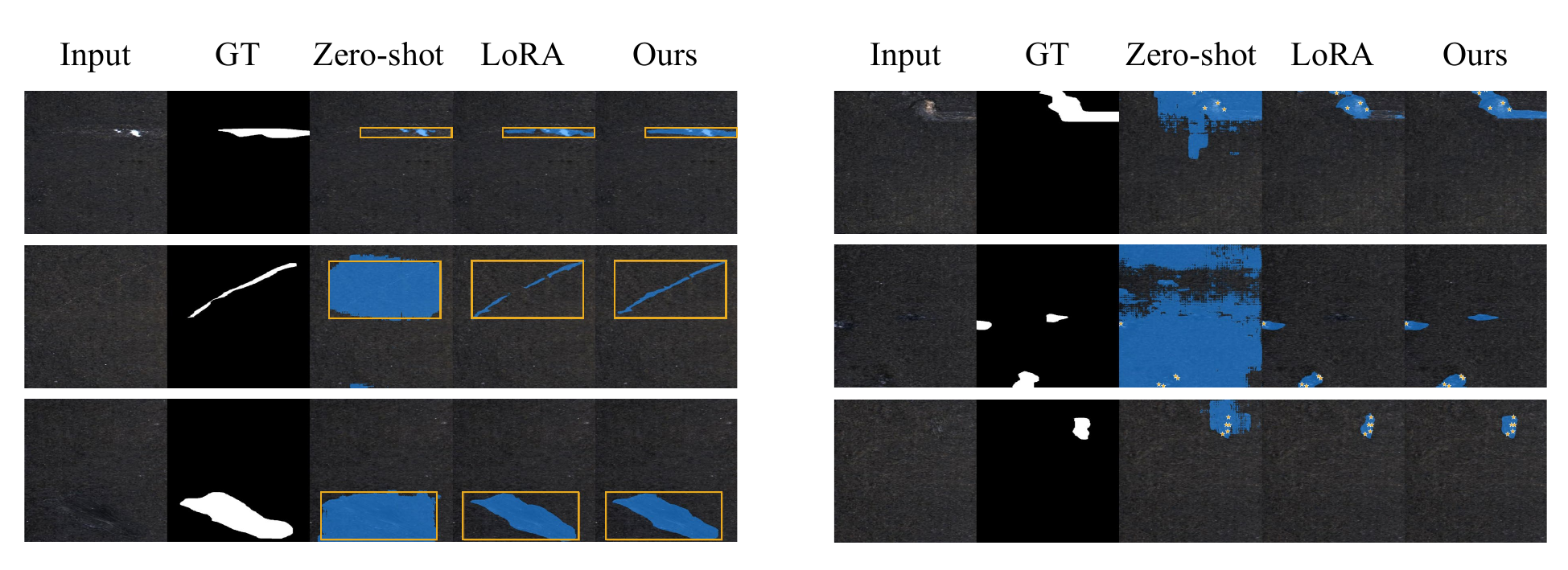}
        \caption{Qualitative analysis on KSDD2 using different prompts. We provide examples with box-level prompts (left) and point-level prompts (right). GT means ground truth.}
        \label{fig:vis_ksdd2}
\end{figure*}

\begin{figure*}[ht!]
\includegraphics[width=0.93\textwidth]{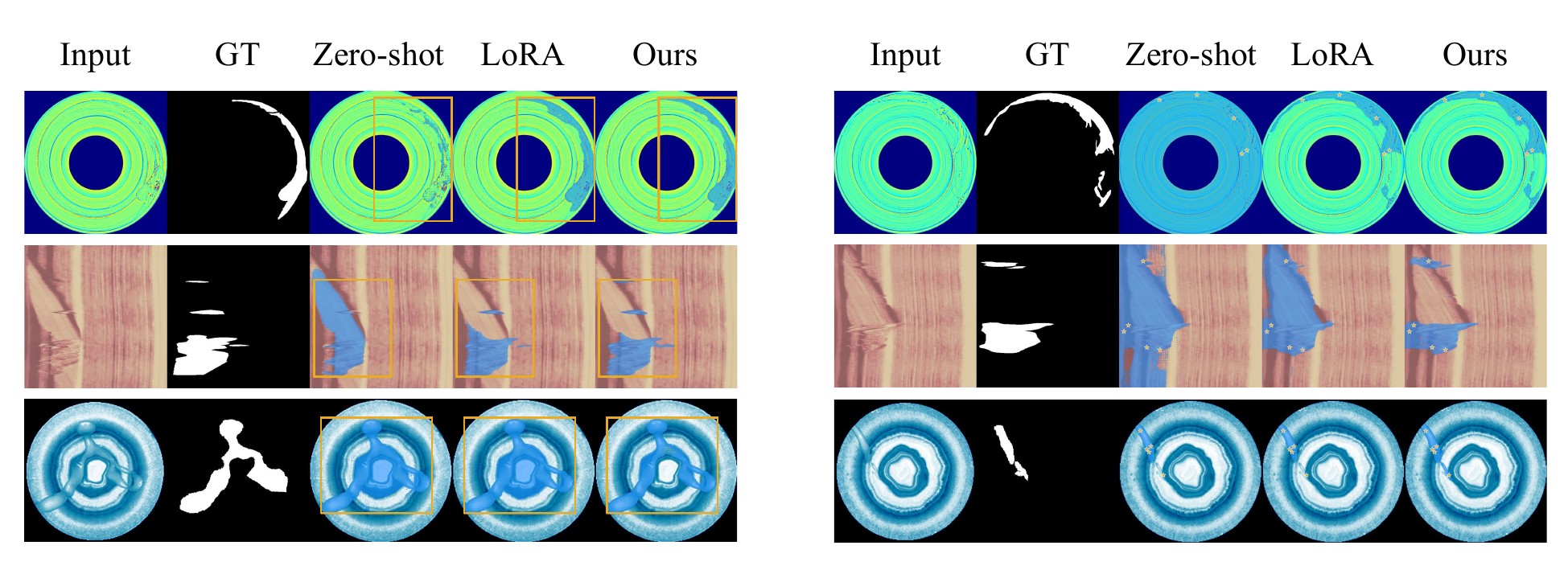}
        \caption{Qualitative analysis on BTAD using different prompts. We provide examples with box-level prompts (left) and point-level prompts (right). GT means ground truth.}
        \label{fig:vis_btad}
    \centering
\end{figure*}

\begin{figure*}[ht!]
\includegraphics[width=0.93\textwidth]{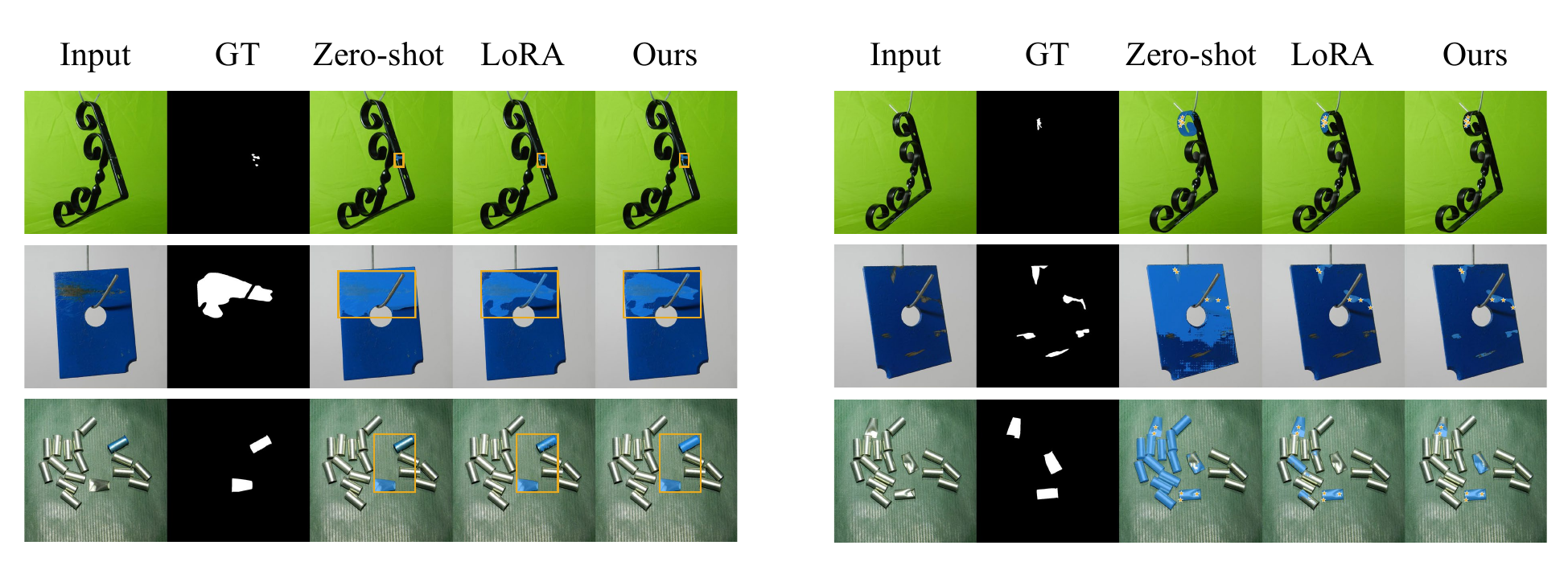}
        \caption{Qualitative analysis on MPDD using different prompts. We provide examples with box-level prompts (left) and point-level prompts (right). GT means ground truth.}
        \label{fig:vis_mpdd}
    \centering
\end{figure*}

\subsection{Visualization of Anomaly Segmentation on Benchmark Datasets}
We provide additional visualization results on VisA (Fig. \ref{fig:vis_visa}), MTD (Fig. \ref{fig:vis_mtd}), KSDD2 (Fig. \ref{fig:vis_ksdd2}), BTAD (Fig. \ref{fig:vis_btad}), and MPDD (Fig. \ref{fig:vis_mpdd}), respectively. We can observe that our method can provide more accurate anomaly segmentation results across different scenarios, compared with the original SAM and the LoRA baseline, well demonstrating the effectiveness and remarkable generalization of the proposed methods.

\end{document}